\documentclass[11pt]{article}

\usepackage[preprint]{neurips_2019}

\usepackage{arxiv}
\usepackage[numbers]{natbib}
\usepackage{hyperref}
\usepackage{cleveref}
\usepackage[utf8]{inputenc} 
\usepackage[T1]{fontenc}    
\usepackage{hyperref}       
\usepackage{url}            
\usepackage{booktabs}       
\usepackage{amsfonts}       
\usepackage{nicefrac}       
\usepackage{microtype}      
\usepackage{graphicx,wrapfig}
\usepackage{amsmath}
\setcitestyle{square}
\title{Poly-GAN: Multi-Conditioned GAN for Fashion Synthesis}

\author{
  Nilesh Pandey, Andreas Savakis
 \thanks{Use footnote for providing further information about author (webpage, alternative address)---\emph{not} for acknowledging
    funding agencies.} \\
  Department of Computer Engineering\\
 Rochester Institute of Technology\\
  Rochester, NY 14623 \\
  \texttt{ np9207@rit.edu andreas.savakis@rit.edu} \\
}

\begin{document}

\maketitle

\begin{abstract}
We present Poly-GAN, a novel conditional GAN architecture that is motivated by Fashion Synthesis, an application where garments are automatically placed on images of human models at an arbitrary pose.
Poly-GAN allows conditioning on multiple inputs and is suitable for many tasks, including image alignment, image stitching and inpainting. 
Existing methods have a similar pipeline where three different networks are used to first align garments with the human pose, then perform stitching of the aligned garment and finally refine the results.
Poly-GAN is the first instance where a common architecture is used to perform all three tasks.
Our novel architecture enforces the conditions at all layers of the encoder and utilizes skip connections from the coarse layers of the encoder to the respective layers of the decoder. Poly-GAN is able to perform a spatial transformation of the garment based on the RGB skeleton of the model at an arbitrary pose.  Additionally, Poly-GAN can perform image stitching, regardless of the garment orientation,
and
inpainting on the garment mask when it contains irregular holes.
Our system achieves state-of-the-art quantitative results on Structural Similarity Index metric and Inception Score metric using the DeepFashion dataset.

\end{abstract}

\section{Introduction}
\label{headings}
Generative Adversarial Networks (GANs) have been one of the most exciting developments in recent years, as they have demonstrated impressive results in various applications including Fashion Synthesis \citet{Prada}, \citet{StyleGAN}, \citet{VITON}.

Fashion Synthesis is a challenging task that requires placing a reference garment on a source model who is at an arbitrary pose and wears a different garment \citet{VITON} \citet{CP-VTON} \citet{MG-VTON} \citet{ClothingGAN} \citet{FashionGAN2}. The arbitrary human pose requirement creates challenges, such as handling self occlusion or limited availability of training data, as the training dataset may or not have the model's desired pose. Some of the challenges in Fashion Synthesis are encountered in other applications, such as person re-identification \citet{SoftGAN}, person modeling  \citet{P-rid}, and Image2Image translation \citet{CanalogyGAN}. 
Existing methods for Fashion Synthesis follow a pipeline consisting of three stages, each requiring different tasks, that are performed by different networks.  These tasks include performing an affine transformation to align the reference garment with the source model \citet{GMM}, stitching the garment on the source model, and refining or post-processing to reduce artifacts after stitching. The problem encountered with this pipeline is that stitching the warped garment often results in artifacts due to self occlusion, spill of color, and blurriness in the generation of missing body regions.

In this paper, we take a more universal approach by proposing a single architecture for all three tasks in the Fashion Synthesis pipeline.  
Instead of using an affine transformation to warp the garments to the body shape, we generate garments with our GAN conditioned on an arbitrary human pose. Generating transformed garments overcomes the problem of self occlusion and  generates occluding arms and other body parts very effectively. The same architecture is then trained to perform stitching and inpainting.  
We demonstrate that our proposed GAN architecture not only achieves state of the art results for Fashion Synthesis, but it is also suitable for many tasks. Thus, we name our architecture Poly-GAN. 
Figure 1  shows representative examples of the performance achieved with Poly-GAN.


Our Fashion Synthesis approach consists of the following stages, illustrated in Figure \ref{fig:2}. 
Stage 1 performs image generation conditioned on an arbitrary human pose, which  changes the shape of the reference garment so it can precisely fit on the human body. 
Stage 2 performs image stitching of the newly generated garment (from Stage 1) with the model after the original garment is segmented out. 
Stage 3 performs refinement by inpainting the output of Stage 2 to fill any missing regions or spots. Stage 4 is a post-processing step that combines the results from Stages 2 and 3, and adds the model head for the final result. Our approach achieves state of the art quantitative results compared to the popular Virtual Try On (VTON) method \citet{CP-VTON}.

The main contributions of this paper can be summarized as follows:
\begin{enumerate}
\vspace{-0.12in}
\item We propose a new conditional GAN architecture, which can operate on multiple conditions that manipulate the generated image. 
\vspace{-0.015in}
\item In our Poly-GAN architecture, the conditions are fed to all layers of the encoder to strengthen their effects throughout the encoding process.  Additionally, skip connections are introduced from the coarse layers of the encoder to the respective layers of the decoder.
\vspace{-0.015in}
\item We demonstrate that our architecture can perform many tasks, including shape manipulation conditioned on human pose for affine transformations, image stitching of a garment on the model, and image inpainting.
\vspace{-0.015in}
\item Poly-GAN is the first GAN to perform  an affine transformation of the reference garment based on the RGB skeleton of the model at an arbitrary pose. 
\vspace{-0.01in}
\item  Our method is able to preserve the desired pose of human arms and hands without color spill, even in cases of self occlusion, while performing Fashion Synthesis.   
\end{enumerate}{}
        


\begin{figure}
 \begin{center}
  \includegraphics[width=0.6\linewidth]{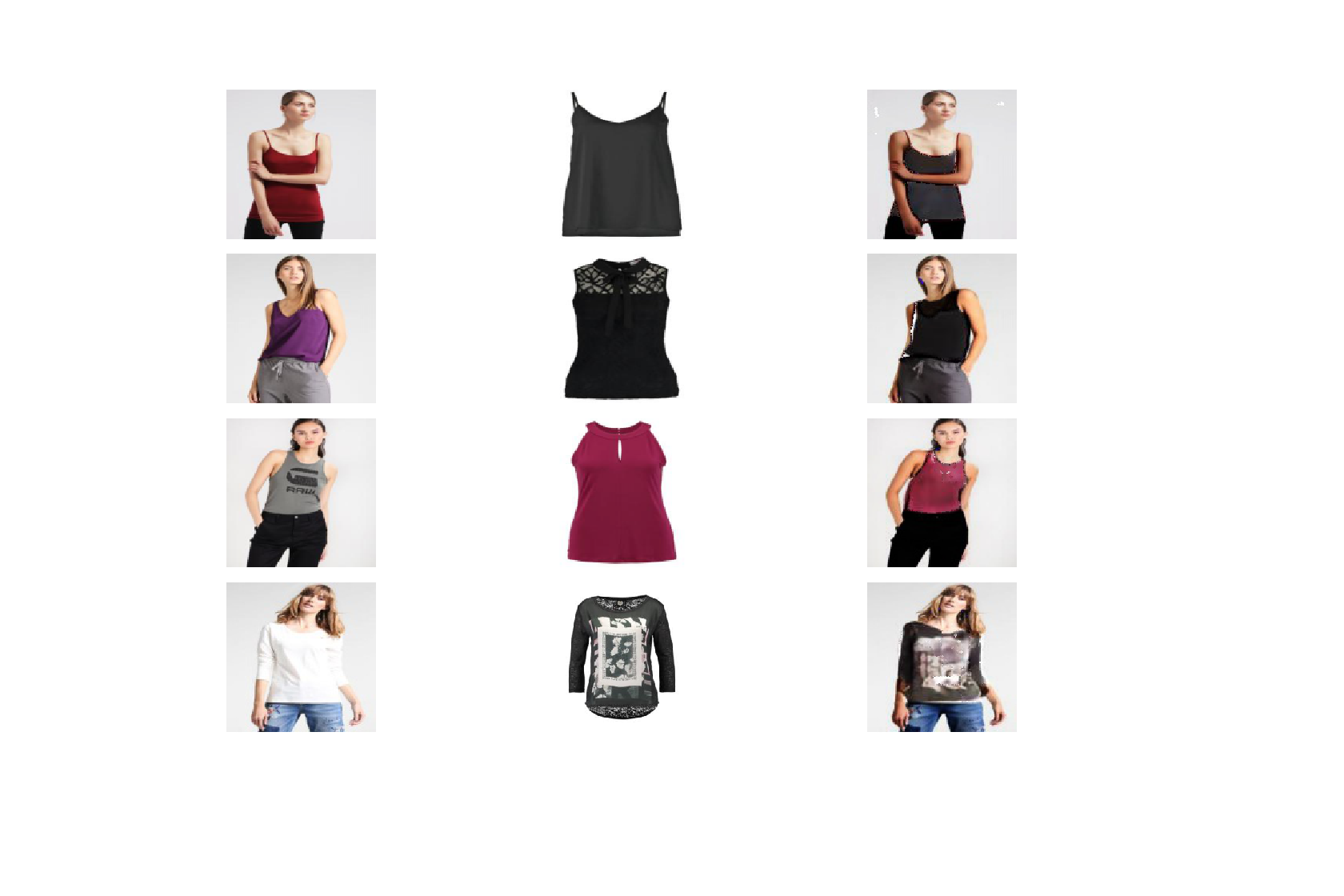}
  \caption{Examples of Fashion Synthesis results generated with Poly-GAN. Shown in the columns from left to right: model image, reference garment, Poly-GAN result.}
  \label{fig:1}
  \end{center}
\end{figure}

\section{Related Work}
Building on the successes of generative adversarial nets \citet{GAN}, 
conditional GANs \citet{ConGAN},\citet{InfoGAN} incorporate a specific conditional restriction 
in their generator network, so that it learns to generate fake samples under that condition. 
Conditional GAN incorporate a binary mask or label as conditional input by concatenating it with the input image or with a latent noise vector.

In the progressive GAN architecture
\citet{PGAN}, layers are added to both the generator and the discriminator during training. 
The addition of new layers increases the fine detail as training progresses. 
A random noise vector is given as an input to the generator. It is possible to have an embedding layer \citet{ConGAN} at the input of the Progressive GAN which provides class conditional information to the network. However, even with the class information, the generated image will be random in nature with no control over the generated structure or texture of the image.

The family of methods that try to swap the source garment with the target garment are referred to as Virtual Try On Networks (VTON) \citet{CP-VTON} \citet{MG-VTON} \citet{VITON}. Methods that fall under VTON usually have a similar pipeline consisting of three main components performed by different networks: a) Pose Alignment Network b) Stitch/Swap Network c) Refinement Network. The pose alignment network aligns the target garment with the source image by learning an affine transformation \citet{GMM}. The stitch/swap network is a GAN with some approaches \citet{MG-VTON}, but there are cases where it simply performs stitching \citet{CP-VTON}.
The refinement process generally differs among methods, as every approach has different shortcomings to refine.

The GAN based VTON methods \citet{CanalogyGAN}, \citet{FashionGAN2}, \citet{ClothingGAN} are becoming popular due to the ability of their GAN to generate images conditioned on image data such as mask, garment or pose.  
One of the first methods to use GANs for VTON \citet{CanalogyGAN} uses the cycle consistency approach to generate humans in different garments. The conditional GAN in \citet{CanalogyGAN} can accept the reference garment, model image, and model garment as inputs to the network, which gives the ability to control the generation of results conditioned on the reference garment.


\section{Poly-GAN}
\label{headings2}
\begin{figure}
\centering
  \includegraphics[width=\linewidth]{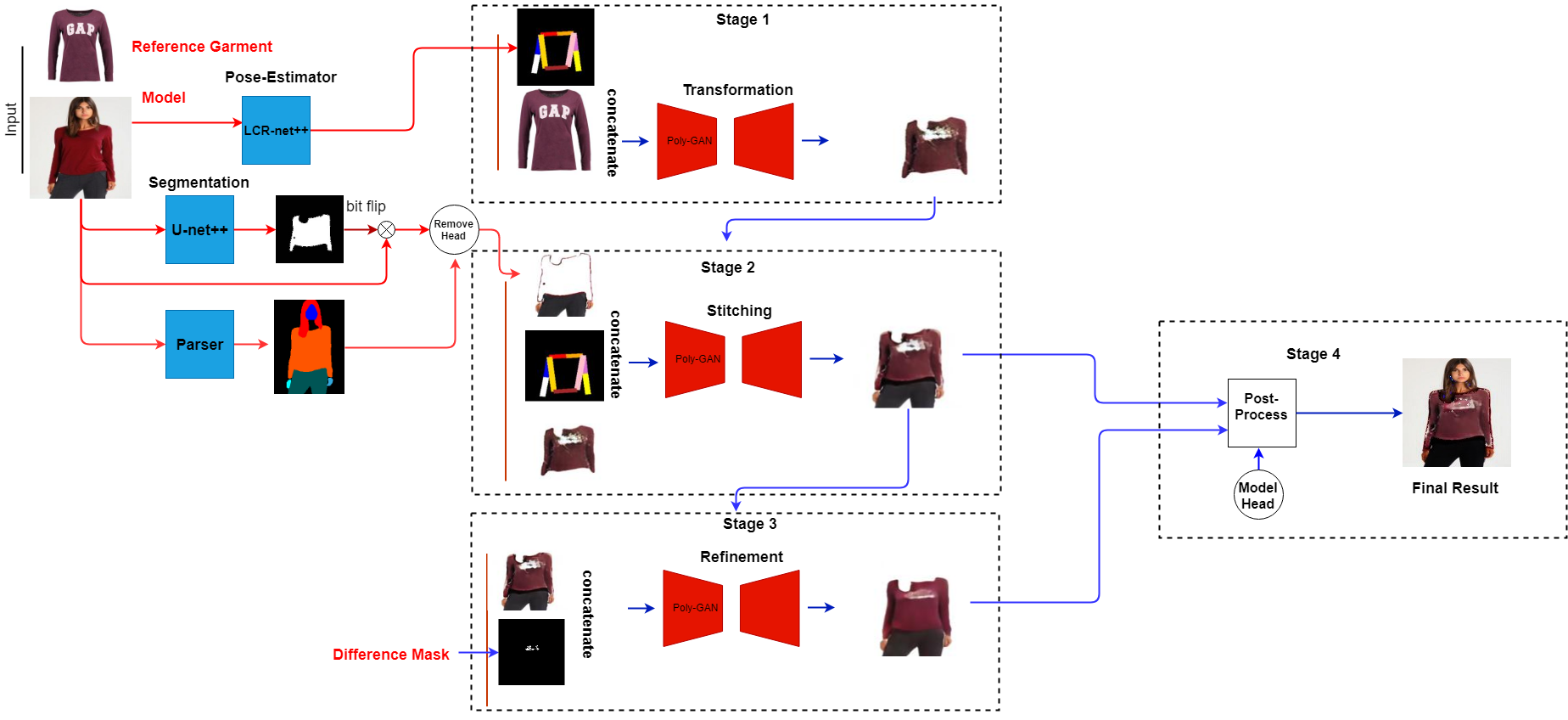}
  \caption{Poly-GAN pipeline. Stage 1: Garment transformation with Poly-GAN conditioned on the RGB skeleton of the model and the reference garment. Stage 2: Garment stitching with Poly-GAN conditioned on the segmented model, the RGB skeleton and the transformed garment. Stage 3: Refinement for hole filling with Poly-GAN conditioned on the stitched image and difference mask indicating missing regions. Stage 4: Postprocessing  for combining the outputs of Stages 2 and 3 with the model head for the final result.}
  \label{fig:2}
\end{figure}
Poly-GAN, a new conditional GAN for fashion synthesis, is the first instance where a common architecture is used to perform many tasks previously performed by different networks. Poly-GAN is flexible and can accept multiple conditions as inputs for various tasks. We begin with an overview of the pipeline used for Fashion Synthesis, shown in Figure \ref{fig:2}, and then we present the details of the Poly-GAN architecture. 
\subsection{Pipeline for Fashion Synthesis}
Two images are inputs to the pipeline, namely the reference garment image and the model image, which is the source person on whom we wish to place the reference garment. 
A pre-trained pose estimator is used to extract the pose skeleton of the model, as shown in Figure \ref{fig:2}. The model image is passed to the segmentation network to extract the segmented mask of the garment, which is used to replace the old garment on the model.

The entire flow can be divided into 4 stages illustrated in Figure \ref{fig:2}. 
In Stage 1, the RGB pose skeleton is concatenated with the reference garment and passed to the Poly-GAN. The RGB skeleton acts as a condition for generating a garment that is reshaped according to an arbitrary human pose. 
The Stage 1 output is a newly generated garment 
which matches the shape and alignment of the RGB skeleton on which Poly-GAN is conditioned. 
The transformed garment from Stage 1 along with the segmented human body
(without garment and without head) and the RGB skeleton are passed to the Poly-GAN in Stage 2. 
Stage 2 serves the purpose of stitching the generated garment from Stage 1 to the segmented human body which has no garment. In Stage 2, Poly-GAN assumes that the incoming garment may be positioned at any angle and does not require garment alignment with the human body. 
This assumption makes Poly-GAN more robust to potential misalignment of the generated garment during the transformation in Stage 1. 
Due to differences in size between the reference garment and the segmented garment on the body of the model, there may be blank areas due to missing regions.
To deal with missing regions at the output of Stage 2, we pass the resulting image to Stage 3 along with the difference mask indicating missing regions.
In Stage 3, Poly-GAN learns to perform inpainting on irregular holes and refines the final result. 
In Stage 4, we perform post processing
by combining the results of Stage 2 and Stage 3, and stitching the head back on the body for the final result. 


We used established architectures for segmentation and pose estimation. For the garment segmentation task, we trained U-Net++ \citet{Unet++} \citet{Unet} from scratch on the DeepFashion dataset \citet{DeepFashion}.
It is important for the segmentation module to precisely separate the source garment from the body, so that the reference garment can be placed at the correct location. 
Otherwise the segmentation process could lead to artifacts in the form of holes or a visible outline around the garment. These artifacts were sometimes present in the U-Net++ segmentation results.

To extract the RGB skeleton, we used the pretrained LCR-net++ pose estimation method \citet{LCR}.
The network was trained on MPII Human Pose dataset \citet{MPI2} and performed reliably. The pose estimator generated a missing pose even in cases with partial occlusion. 

We also created human parsing data using a pre-trained method \citet{MULA} \citet{Pytorch-MULA}. The human parsing data is used for segmenting the head from the body of the model, as well as for creating data to train U-Net++.

\subsection{Poly-GAN Architecture}
\begin{figure}
\centering
  \includegraphics[width=\columnwidth]{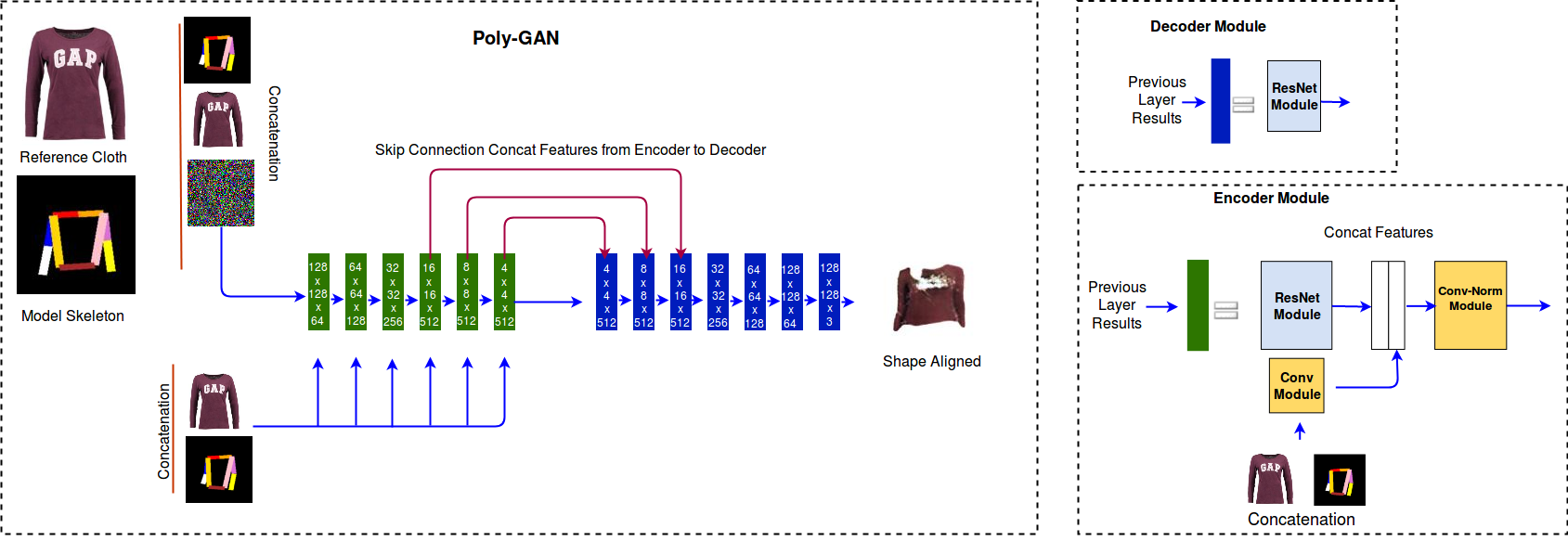}
  \caption{Poly-GAN architecture. The left side shows the encoder in green and decoder in blue. The example shown is for garment transformation (Stage 1) but the architecture is the same for image stitching and inpainting. The conditions of reference garment and body pose are fed at all of the encoder layers. Skip connections from coarse layers of the encoder are fed to the corresponding layers of the decoder. The top right block shows the decoder module and the bottom right block shows the encoder module. }
  \label{fig:4}
\end{figure}

\begin{figure}
\begin{center}
  \includegraphics[width=0.9\linewidth]{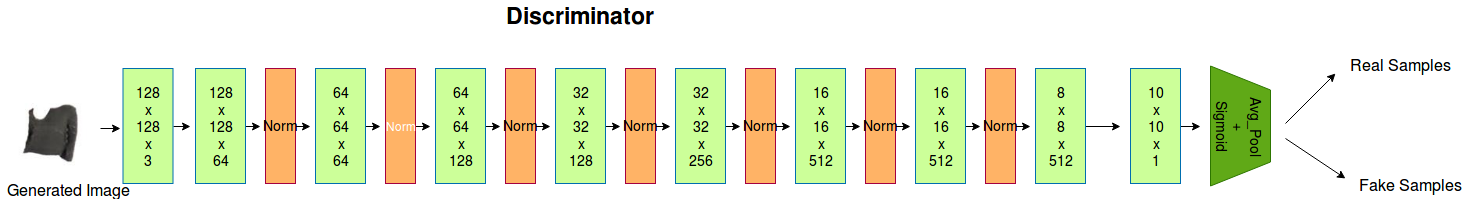}
  \caption{Architecture of the Discriminator. The example shown is for garment transformation but the architecture is the same for image stitching (Stage 2) and inpainting (Stage 3). }
  \label{fig:5}
\end{center}
\end{figure}

The Poly-GAN architecture, shown in Figure \ref{fig:4}, follows the encoder-decoder style of generator.
The discriminator architecture is shown in Figure \ref{fig:5}.
The network architecture is inspired by recent developments in generative adversarial networks, and includes some unique features. We explain the details of our architecture next.

\subsubsection{Encoder}
The encoder in our architecture, shown in Figure \ref{fig:4}, can be divided in three main components, a) Conv Module, b) ResNet Module, and c) Conv-Norm Module. 
We discuss the use of two new modules that are Conv Module and Conv-Norm Module. The ResNet module follows the standard architecture of residual networks \citet{Resnet}.

\textbf{Conv Module.}
Conditional GANs \citet{ConGAN}, \citet{InfoGAN} 
have conditional inputs at the first layer. e.g., the source image or latent noise. 
An important benefit of our architecture is that it includes conditional inputs at every layer, instead of only having conditional inputs at the first layer. 
We observed that if the conditional input is limited to the first layer, then its effect diminishes as the features propagate deeper in the network. As a result, there is hardly any conditional information left in deeper layers. 
We decided to feed the conditional inputs at every layer through a Conv module shown in the encoder module of Figure \ref{fig:4}. 
The Conv Module consists of 3 convolution layers, each followed by a ReLU activation function which outputs the number of features required in the layer under consideration.

\textbf{Conv-Norm Module.}
The ResNet module outputs newly learned features, which are concatenated with the features learned from the Conv Module. There is significant difference between the features from ResNet and features from the Conv Module. 
Thus, we pass the concatenated features to the Conv-Norm module to take advantage of the variation in features from the two modules. 
The Conv-Norm module has 2 convolution layers, with each convolution layer followed by an instance normalization layer and activation function.

\subsubsection{Decoder}
The decoder part of the network learns to generate the desired image based on the features encoded by the encoder. 
The decoder part of Poly-GAN, shown in the decoder block of Figure \ref{fig:4}, is similar to the decoder used in other GANs, e.g., Cycle GAN \citet{CycleGAN}. 
The decoder consists of a series of ResNet modules followed by transposed convolution to up-sample the input in each block.
However, there are fewer layers in the decoder which receive information from the encoder through skip connections.

\subsubsection{Skip Connections}
\begin{figure}
 \begin{center}
 
  \includegraphics[width=4.5in]{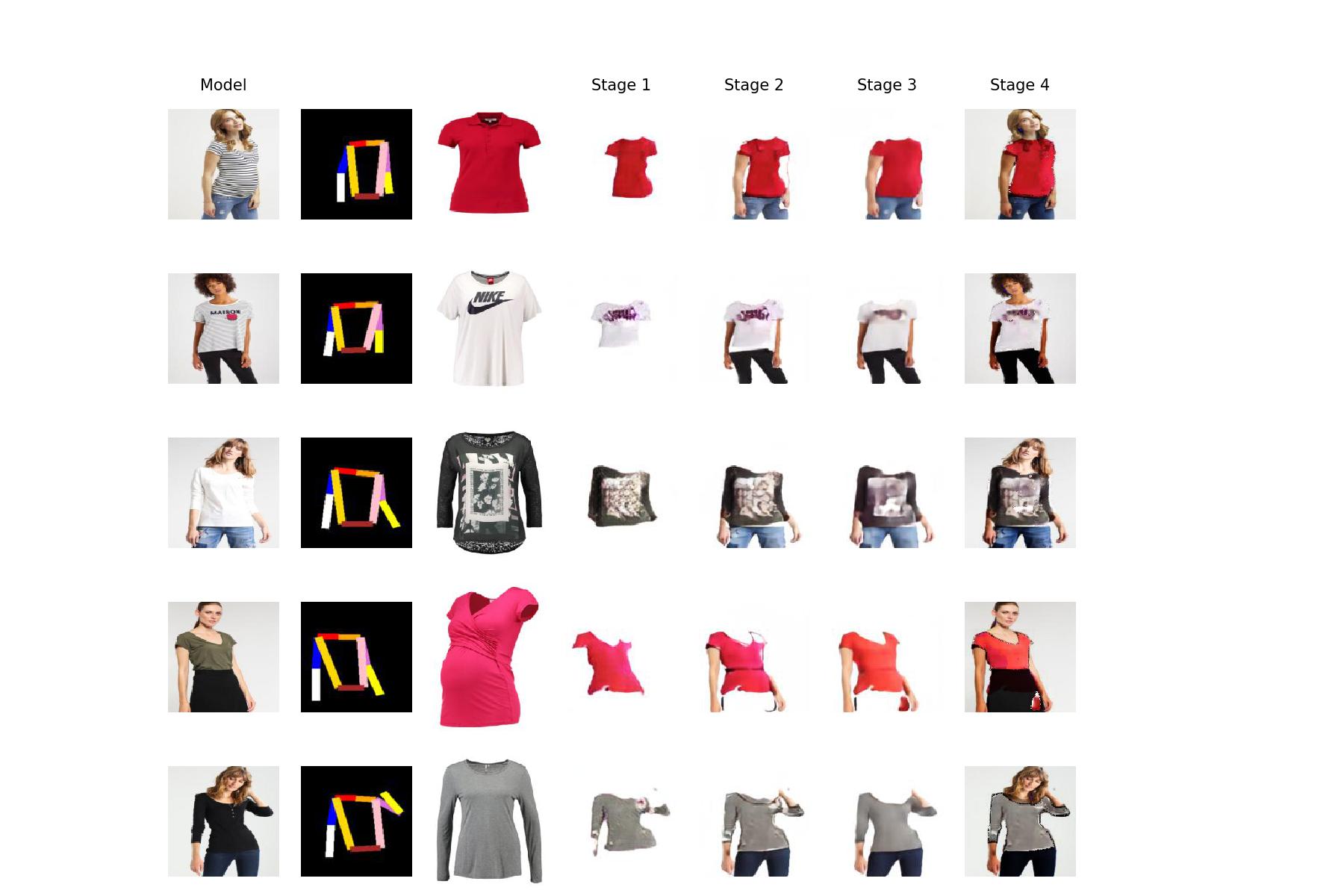}
 \caption{Poly-GAN results shown from left to right column: Model image; Pose Skeleton; Reference Garment; Stage 1: transformed garment; Stage 2: garment stitched on segmented model; Stage 3: refinement of outputs from stage 2: Stage 4: post-process result from stage 2 and stage 3 with head in original model image. }
  \label{fig:6}
 \end{center}
\end{figure}

One of the important design decisions of Poly-GAN is the placement of skip connections. 
From previous works \citet{StyleGAN}, it is understood that coarser spatial resolution results in higher level spatial change, such as pose and shape. 
We use skip connections to pass the information from the encoded coarse layers (4x4, 8x8, 16x16) to the respective decoder layers (4x4, 8x8, 16x16) and concatenate to augment the feature representation. 
We observed that if we use skip connections between encoder and decoder at spatial resolution above (16x16), then the generated image is not deformable enough to be close to the ground truth. If we don't use skip connections above spatial resolution of (16x16), then the problem of missing minute details arises in the generated image. This is illustrated in the results of Figure \ref{fig:6}. 
Using skip connection to connect all layers of the encoder to their respective decoder layer would help in passing minute details from the encoder side, but would also hinder learning and generating new images effectively.

\subsection{Discriminator}
The discriminator used in our experiments is the same discriminator used in the Super Resolution GAN \citet{SRGAN}.
The architecture of the discriminator is shown in Figure \ref{fig:5}.
The reason behind the selection of the discriminator is to penalize for blurry images, which are often generated by GANs using the $L_{2}$ loss function.

\subsection{Loss Function}
Our loss function consists of three components: adversarial loss $L_{adv}$, GAN loss $L_{gan}$ and identity loss $L_{id}$. The total loss and its components are presented below, where $D$ is the discriminator, $G$ is the generator, $x_{i}, i=1,...,N$ represent $N$ input samples from different distributions $p_{i}(x_{i}), i=1,...N$, $t$ is the target, $F$ are fake labels and $R$ are real labels.
\begin{equation}
    \mathit{L_{loss}} = \mathit{L_{adv}} + \mathit{L_{gan}} + \mathit{l_{id}}
    \label{eq:4}
\end{equation}
\begin{equation}
    \underset{D}{\mathit{min}}\mathit{L_{adv}} = \lambda_{1}E_{t\sim p_{d}(t)}{\left \| D(t)-R \right \|_{2}^{2}} + \lambda_{2}E_{x_{1}\sim p_{1}(x_{1}),..,x_{N}\sim p_{N}(x_{N})}{\left \| D(G(x_{1},..,x_{N})-F \right \|_{2}^{2}}
    \label{eq:1}
\end{equation}
\begin{equation}
    \underset{G}{\mathit{min}}\mathit{L_{gan}} = \lambda_{3}E_{x_{1}\sim p_{1}(x_{1}),..,x_{N}\sim p_{N}(x_{N})}{\left \| D(G(x_{1},..,x_{N})-R \right \|_{2}^{2}}
    \label{eq:2}
\end{equation}
\begin{equation}
    L_{id}=\lambda_{4}E_{t\sim p_{d}(t),x_{1}\sim p_{1}(x_{1}),..,x_{N}\sim p_{N}(x_{N})}{\left \| G(x_{1},..,x_{N})-t \right \|_{1}}
    \label{eq:3}
\end{equation}

In the above equations, 
 $\lambda_{1}$ to $\lambda_{4}$ are hyperparamters that are tuned during training.
We use the $L_{2}$ function as the adversarial loss for our GAN, similar to \citet{LSGAN}.
We also use the $L_{1}$ function for the identity loss, which helps reduce texture and color shift between the generated image and the ground truth. 
We considered adding other loss functions, such as the perceptual loss \citet{Perceptual-loss} and SSIM loss \citet{SSIM}, but they did not improve our results.  

\section{Experiments}

\subsection{Training Methodology}
The work on StyleGAN \citet{StyleGAN} revealed certain input conditions that contributed to the generation of super realistic images, improving upon the blurry or distorted images generated by previous works. These conditions are a) images in the dataset should be of similar zoom ratio; b) the input to the GAN should be simple; c) diversity of the data should be balanced.
In our initial experiments with DCGAN \citet{DCGAN}, ProgressiveGAN \citet{PGAN}, CycleGAN \citet{CycleGAN}, and StyleGAN, we observed that if we pass the whole body with the head attached, then the results are blurry and the GAN fails to generate properly. In our experiments we made sure that the dataset is clean enough to have similar zoom ratio on the subject, images are diverse, and we removed any complex distractions, such as the head, before passing to the Poly-GAN inputs.

We chose to use the RGB skeleton over a binary image of the pose. Our decision was based on the observation that color coded feature presentation are more effective. The other important observation we made while experimenting with Poly-GAN is that the results are affected by the order in which the conditions are concatenated and we adjusted the inputs accordingly. We used hyperparameters similar to CycleGAN \citet{CycleGAN} for Poly-GAN training. We used a learning rate of 0.0002, an Adam optimizer with $\beta_{1}=0.5$ and $\beta_{1}=0.999$, batch size of 1 in all our experiments, and image size of 128x128 to feed images in the network. We also used an image buffer during training similar to the one suggested in \citet{Shrivastava}. The image buffer helps in stabilizing the discriminator by keeping a history of a fixed number of generated images which are randomly passed to the discriminator. All the experiments are performed on a workstation with 16GB of RAM using an Nvidia 1080ti GPU and Ubuntu 16.04.

\subsection{Dataset}

We follow the process used in VITON \citet{VITON} while creating our training and testing datasets from the publicly available DeepFashion dataset \citet{DeepFashion}. 
We use LCR-net++ \citet{LCR} for 2D human pose estimation, and Pytorch-MULA \citet{MULA} \citet{Pytorch-MULA} to get the parsing results for the models. The human parser \citet{MULA} is used to create data to train U-Net++ \citet{Unet++}, to segment the garments from the models, and to remove the head from the models.
We have 14,221 training samples, as in the VITON dataset, and 900 paired samples for testing our method.
We note that the authors in VITON \citet{VITON} have used \citet{SSL} for human parsing and \citet{RPose} for 2D human pose estimation. 

\begin{table}[h!]
\begin{center}
    \caption{Fashion Synthesis Quantitative Results. Bold numbers indicate best performance.}
    \label{tab:1}
\begin{tabular}{@{}ccc@{}}
\toprule
\multicolumn{1}{c}{} &   Metric      \\ \cmidrule(r){2-3}
Method     & SSIM & IS  \\ \midrule
CP-VTON &0.6889 & 2.6049     \\
Poly-GAN Stage 2 &0.7174 & \textbf{2.8193}      \\
Poly-GAN Stage 3 & \textbf{0.7369} & 2.6549      \\
Poly-GAN Stage 4 &0.7251 & 2.7904      \\ \bottomrule
\end{tabular}
\end{center}
\end{table}

\subsection{Quantitative Evaluation}
We use the Structural Similarity Index metric (SSIM) \citet{SSIM} and Inception Score metric (IS) \citet{Inception score} which are widely accepted metrics for evaluation of images generated by GANs. SSIM predicts the similarity between two images, where the higher the score the better the network is in generating realistic images. The Inception Score compares the quality of an image to human level grading, and is sensitive to blurring in an image.

We compare our results to the state of art method CP-VTON \citet{CP-VTON}. Since the code for CP-VTON is publicly released \citet{CP-VTON-Github}, we are able to obtain results on our dataset for comparison. 
The state of the art method MG-VTON \citet{MG-VTON} tries to solve the problem in a significantly different way than ours, but unfortunately their code is not available. Therefore, our comparison was limited to CP-VTON.
We perform our evaluation on test data that consist of 900 paired images. 
The results are shown in Table \ref{tab:1}. SSIM and IS scores are included for the final stage of Poly-GAN, as well as for Stages 2 and 3 (after stitching the head of the model). The results illustrate that Poly-GAN outperforms CP-VTON in all cases. By comparing the scores of different stages, we see that the output of Stage 2 is the sharpest and has the highest IS score, but it suffers from holes due to blank regions. The output of Stage 3 has the highest SSIM score, i.e. is the most similar to the desired output, but it suffers from blurriness artifacts. The combination on the two gives a balanced score without scoring the highest for either metric.    

\begin{figure}
 \begin{center}
 \includegraphics[width=4.5in]{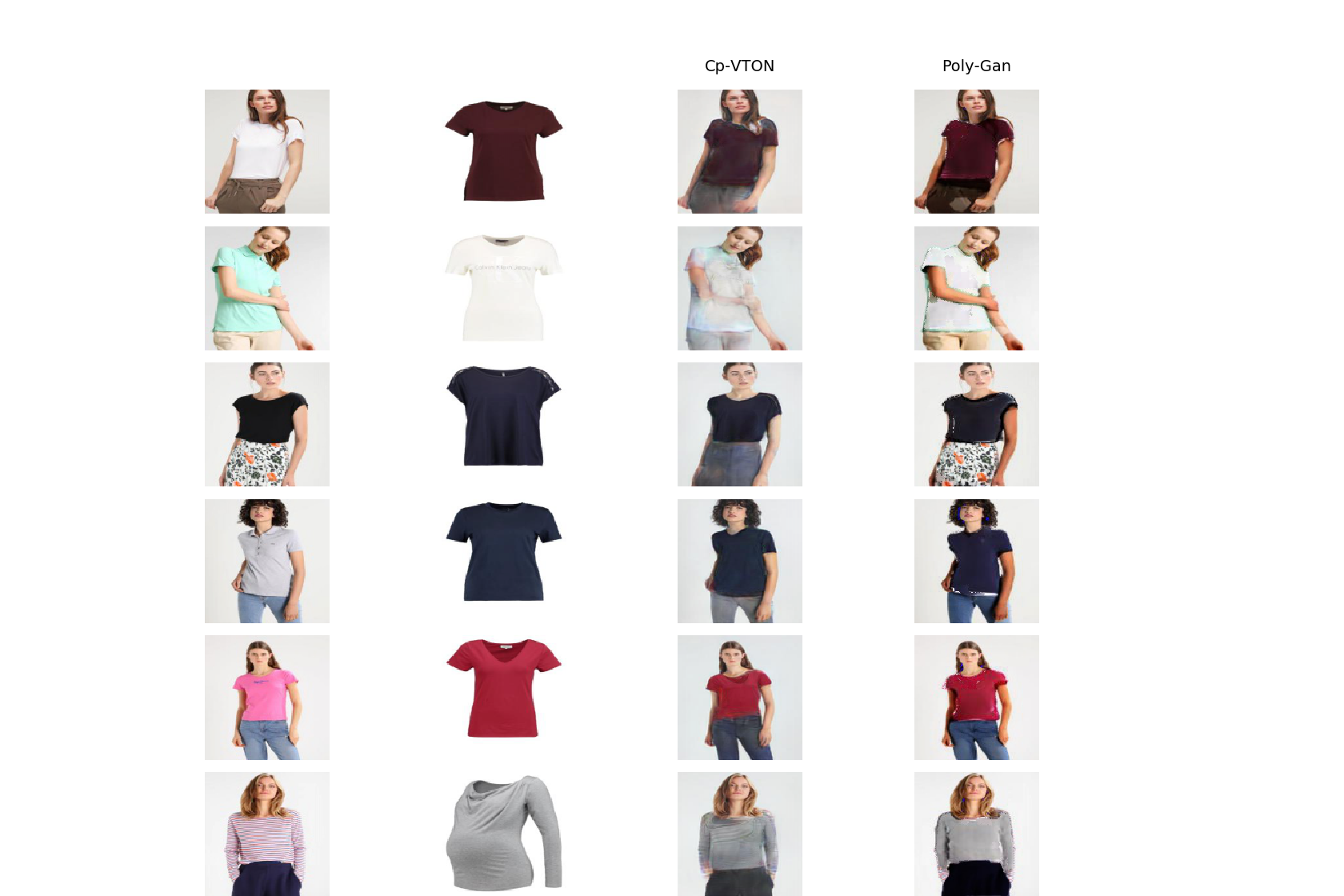}
 \caption{Comparison examples of Poly-GAN with CP-VTON. Shown in columns from left to right: model image, reference garment, CP-VTON result, Poly-GAN result. }
 \label{fig:8}
\end{center}
\end{figure}
\begin{figure}
  \includegraphics[width=\linewidth]{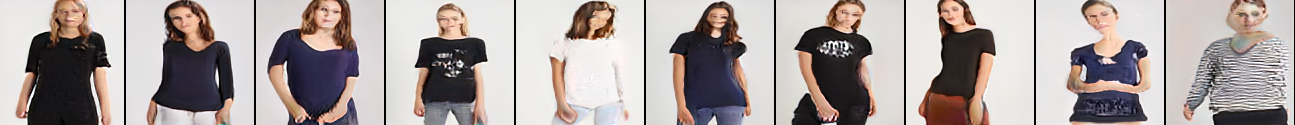}
  \caption{Results of models randomly generated using StyleGAN.}
  \label{fig:7}
\end{figure}
\subsection{Qualitative Evaluation}
We provide a visual comparison of our results with CP-VTON in Figure \ref{fig:8}. 
Our Poly-GAN method is able to keep body parts intact. especially in cases where the arms occlude the body. 
For example, in Figure \ref{fig:8} the second image from top shows that CP-VTON fails in the case of self occlusion which our method is able to handle in almost all cases. Poly-GAN is able to precicely map the generated garment from Stage 1 (see Figure \ref{fig:2}) to the missing garment region on the body in Stage 2, thus avoiding the spill of color, which is evident in CP-VTON.

While the placement of the garment on the model is very good, 
both methods suffer from slight color shift in some samples, 
which is a common problem with GANs that are asked to generate colors from various distributions.
One limitation of our method is texture preservation when generating letters, graphics or patterns that are present in the reference garment. This artifact is due to the use of a GAN to reshape the garment, instead of using image warping methods. A less pronounced artifact in some images is a visible boundary outline that is caused by errors in the segmentation by U-Net++. This limitation can be overcome by using a larger and more diverse dataset for training.

For qualitative evaluation, we also present images that were randomly generated with StyleGAN in Figure \ref{fig:7}. 
It is evident that although StyleGAN is not conditioned on pose, the generated images suffer from color spill due to occlusion and artifacts in preserving textures and graphics on the garments.



\section{Conclusion}
We offer a novel approach to the Fashion Synthesis problem by introducing Poly-GAN, a new multi-conditioned GAN architecture that is suitable for many tasks. Qualitative and quantitative results on DeepFashion demonstrate the benefits of our approach by achieving state of the art results. Future research will focus on further exploring the Poly-GAN architecture for a variety of applications.



\medskip
\begin{thebibliography}{34}

\bibitem[Andriluka et al.(2014)]{MPI2}
Andriluka, M., Pishchulin, L., Gehler, P., \& Schiele, B. (2014). 2D Human Pose Estimation: New Benchmark and State of The Art Analysis. In Proceedings of the IEEE Conference on Computer Vision and Pattern Recognition (pp. 3686-3693).

 \bibitem[Cao et al.(2017)]{RPose}
Cao, Z., Simon, T., Wei, S. E., \& Sheikh, Y. (2017). Realtime Multi-Person 2D Pose Estimation using Part Affinity Fields. In Proceedings of the IEEE Conference on Computer Vision and Pattern Recognition (pp. 7291-7299).

\bibitem[Chen et al.(2016)]{InfoGAN}
Chen, X., Duan, Y., Houthooft, R., Schulman, J., Sutskever, I., \& Abbeel, P. (2016). InfoGAN: Interpretable Representation Learning by Information Maximizing Generative Adversarial Nets. In Advances in Neural Information Processing Systems(pp. 2172-2180).


\bibitem[Cui et al.(2018)]{FashionGAN2}
Cui, Y. R., Liu, Q., Gao, C. Y., \& Su, Z. (2018). FashionGAN: Display your fashion design using Conditional Generative Adversarial Nets. In Computer Graphics Forum (Vol. 37, No. 7, pp. 109-119).


\bibitem[Dong et al.(2019)]{MG-VTON}
Dong, H., Liang, X., Wang, B., Lai, H., Zhu, J., \& Yin, J. (2019). Towards Multi-pose Guided Virtual Try-on Network. arXiv preprint arXiv:1902.11026.
\bibitem[Dong et al.(2018)]{SoftGAN}
Dong, H., Liang, X., Gong, K., Lai, H., Zhu, J., \& Yin, J. (2018). Soft-Gated Warping-GAN for Pose-Guided Person Image Synthesis. In Advances in Neural Information Processing Systems (pp. 474-484).

\bibitem[Gong et al.(2017)]{SSL}
Gong, K., Liang, X., Zhang, D., Shen, X., \& Lin, L. (2017). Look into Person: Self-supervised Structure-sensitive Learning and a New Benchmark for Human Parsing. In Proceedings of the IEEE Conference on Computer Vision and Pattern Recognition (pp. 932-940).

\bibitem[Goodfellow et al.(2014)]{GAN}
Goodfellow, I., Pouget-Abadie, J., Mirza, M., Xu, B., Warde-Farley, D., Ozair, S., ... \& Bengio, Y. (2014). Generative Adversarial Nets. In Advances in Neural Information Processing Systems (pp. 2672-2680).

\bibitem[Han et al.(2018)]{VITON}
Han, X., Wu, Z., Wu, Z., Yu, R., \& Davis, L. S. (2018). VITON: An Image-based Virtual Try-on Network. In Proceedings of the IEEE Conference on Computer Vision and Pattern Recognition (pp. 7543-7552).

\bibitem[He, K. et al.(2016)]{Resnet}
He, K., Zhang, X., Ren, S., \& Sun, J. (2016). Deep Residual Learning for Image Recognition. In Proceedings of the IEEE Conference on Computer Vision and Pattern Recognition (pp. 770-778).

\bibitem[Jetchev et al.(2017)]{CanalogyGAN}
Jetchev, N., \& Bergmann, U. (2017). The Conditional Analogy GAN: Swapping Fashion Articles on People Images. In Proceedings of the IEEE International Conference on Computer Vision (pp. 2287-2292).

\bibitem[Johnson et al.(2016)]{Perceptual-loss}
Johnson, J., Alahi, A., \& Fei-Fei, L. (2016). Perceptual Losses for Real-Time Style Transfer and Super-Resolution. In European Conference on Computer Vision (pp. 694-711). Springer, Cham.

\bibitem[Karras et al.(2017)]{PGAN}
Karras, T., Aila, T., Laine, S., \& Lehtinen, J. (2017). Progressive Growing of GANs for Improved Quality, Stability, and Variation. arXiv preprint arXiv:1710.10196.

\bibitem[Karras et al.(2018)]{StyleGAN}
Karras, T., Laine, S., \& Aila, T. (2018). A Style-Based Generator Architecture for Generative Adversarial Networks. arXiv preprint arXiv:1812.04948.

\bibitem[Lassner et al.(2017)]{ClothingGAN}
Lassner, C., Pons-Moll, G., \& Gehler, P. V. (2017). A Generative Model of People in Clothing. In Proceedings of the IEEE International Conference on Computer Vision (pp. 853-862).

\bibitem[Ledig et al.(2017)]{SRGAN}
Ledig, C., Theis, L., Huszár, F., Caballero, J., Cunningham, A., Acosta, A., ... \& Shi, W. (2017). Photo-Realistic Single Image Super-Resolution using a Generative Adversarial Network. In Proceedings of the IEEE Conference on Computer Vision and Pattern Recognition (pp. 4681-4690).

\bibitem[Liu et al.(2016)]{DeepFashion}
Liu, Z., Luo, P., Qiu, S., Wang, X., \& Tang, X. (2016). Deepfashion: Powering Robust Clothes Recognition and Retrieval with Rich Annotations. In Proceedings of the IEEE Conference on Computer Vision and Pattern Recognition (pp. 1096-1104).

\bibitem[LiuJ et al.(2018)]{P-rid}
 Liu, J., Ni, B., Yan, Y., Zhou, P., Cheng, S., \& Hu, J. (2018). Pose Transferrable Person Re-Identification. In Proceedings of the IEEE Conference on Computer Vision and Pattern Recognition (pp. 4099-4108).

\bibitem[Mao et al.(2017)]{LSGAN}
 Mao, X., Li, Q., Xie, H., Lau, R. Y., Wang, Z., \& Paul Smolley, S. (2017). Least Squares Generative Adversarial Networks. In Proceedings of the IEEE International Conference on Computer Vision (pp. 2794-2802).

\bibitem[ Mirza et al.(2014)]{ConGAN}
 Mirza, M., \& Osindero, S. (2014). Conditional Generative Adversarial Nets.
 
\bibitem[Nie et al.(2018)]{MULA}
Nie, X., Feng, J., \& Yan, S. (2018). Mutual Learning to Adapt for Joint Human Parsing and Pose Estimation. In Proceedings of the European Conference on Computer Vision (ECCV) (pp. 502-517).

\bibitem[Nie et al.(2018)]{Pytorch-MULA}
Nie, X. (2018). Mutual Learning to Adapt for Joint Human Parsing and Pose Estimation. Retrieved from
https://github.com/NieXC/pytorch-mula

\bibitem[Radford et al.(2015)]{DCGAN}
Radford, A., Metz, L., \& Chintala, S. (2015). Unsupervised Representation Learning with Deep Convolutional Generative Adversarial Networks. arXiv preprint arXiv:1511.06434.

\bibitem[Rogez et al.(2019)]{LCR}
Rogez, G., Weinzaepfel, P., \& Schmid, C. (2019). LCR-net++: Multi-Person 2d and 3d Pose Detection in Natural Images. IEEE Transactions on Pattern Analysis and Machine Intelligence.

\bibitem[Rocco et al.(2017)]{GMM}
Rocco, I., Arandjelovic, R., \& Sivic, J. (2017). Convolutional Neural Network Architecture for Geometric Matching. In Proceedings of the IEEE Conference on Computer Vision and Pattern Recognition (pp. 6148-6157).

\bibitem[Ronneberger et al.(2015)]{Unet}
Ronneberger, O., Fischer, P., \& Brox, T. (2015). U-Net: Convolutional Networks for Biomedical Image Segmentation. In International Conference on Medical Image Computing and Computer-Assisted Intervention (pp. 234-241). Springer, Cham.

\bibitem[Salimans et al.(2016)]{Inception score}
Salimans, T., Goodfellow, I., Zaremba, W., Cheung, V., Radford, A., \& Chen, X. (2016). Improved Techniques for Training GANs. In Advances in Neural Information Processing Systems (pp. 2234-2242).

\bibitem[Shrivastava et al.(2017)]{Shrivastava}
Shrivastava, A., Pfister, T., Tuzel, O., Susskind, J., Wang, W., \& Webb, R. (2017). Learning from Simulated and Unsupervised Images through Adversarial Training. In Proceedings of the IEEE Conference on Computer Vision and Pattern Recognition (pp. 2107-2116).

\bibitem[WangB et al.(2018)]{CP-VTON}
Wang, B., Zheng, H., Liang, X., Chen, Y., Lin, L., \& Yang, M. (2018). Toward Characteristic-Preserving Image-based Virtual Try-On Network. In Proceedings of the European Conference on Computer Vision (ECCV) (pp. 589-604).

\bibitem[WangB et al.(2018)]{CP-VTON-Github}
Wang, B. (2018). CP-VTON
 - Reimplemented code for "Toward Characteristic-Preserving Image-based Virtual Try-On Network". Retrieved from https://github.com/sergeywong/cp-vton.

\bibitem[WangZ et al.(2004)]{SSIM}
 Z. Wang, A. C. Bovik, H. R. Sheikh and E. P. Simoncelli. (2004). Image Quality Assessment: From Error Visibility to Structural Similarity. IEEE Transactions on Image Processing, vol. 13, no. 4 (pp. 600-612).
\bibitem[ZhouZ et al.(2018)]{Unet++}
Zhou, Z., Siddiquee, M. M. R., Tajbakhsh, N., \& Liang, J. (2018). UNet++: A Nested U-Net Architecture for Medical Image Segmentation. In Deep Learning in Medical Image Analysis and Multimodal Learning for Clinical Decision Support (pp. 3-11). Springer, Cham.

\bibitem[ZhuS et al.(2017)]{Prada}
Zhu, S., Fidler, S., Urtasun, R., Lin, D.,  \& Loy, C. C. (2017). Be Your Own Prada: Fashion Synthesis with Structural Coherence. In Proceedings of the IEEE International Conference on Computer Vision (ICCV).

\bibitem[ZhuJ et al.(2017)]{CycleGAN}
Zhu, J. Y., Park, T., Isola, P., \& Efros, A. A. (2017). Unpaired Image-to-Image Translation using Cycle-Consistent Adversarial Networks. In Proceedings of the IEEE International Conference on Computer Vision (pp. 2223-2232).



\end{thebibliography}
\end{document}